\documentclass[pmlr]{jmlr}

\usepackage{longtable}
\usepackage{caption}
\usepackage{subcaption}
\usepackage{adjustbox}
\usepackage{comment}
\usepackage{float}
\usepackage{booktabs}
\usepackage[load-configurations=version-1]{siunitx} 

\theorembodyfont{\upshape}
\theoremheaderfont{\scshape}
\theorempostheader{:}
\theoremsep{\newline}

\jmlrvolume{85}
\jmlryear{2018}
\jmlrworkshop{Machine Learning for Healthcare}

\title[Multiple Instance Learning for ECG Risk Stratification]{Multiple Instance Learning for ECG Risk Stratification}

\author{\Name{Divya Shanmugam} \Email{divyas@mit.edu} 
       \addr \\
       Massachusetts Institute of Technology\\
       Cambridge, MA, USA
       \AND
       \Name{Davis Blalock} \Email{dblalock@mit.edu} 
       \addr \\
       Massachusetts Institute of Technology\\
       Cambridge, MA, USA
       \AND
       \Name{John Guttag} \Email{guttag@mit.edu} 
       \addr\\
       Massachusetts Institute of Technology\\
       Cambridge, MA, USA}
\begin{document}

\maketitle



\begin{abstract}
Patients who suffer an acute coronary syndrome are at elevated risk for adverse cardiovascular events such as myocardial infarction and cardiovascular death. Accurate assessment of this risk is crucial to their course of care. We focus on estimating a patient's risk of cardiovascular death after an acute coronary syndrome based on a patient's raw electrocardiogram (ECG) signal. Learning from this signal is challenging for two reasons: 1) positive examples signifying a downstream cardiovascular event are scarce, causing drastic class imbalance, and 2) each patient's ECG signal consists of thousands of heartbeats, accompanied by a single label for the downstream outcome. Machine learning has been previously applied to this task, but most approaches rely on hand-crafted features and domain knowledge. We propose a method that learns a representation from the raw ECG signal by using a multiple instance learning framework. We present a learned risk score for cardiovascular death that outperforms existing risk metrics in predicting cardiovascular death within 30, 60, 90, and 365 days on a dataset of 5000 patients.
\end{abstract}

\section{Introduction}

 Machine learning has led to improved risk stratification models for a number of outcomes, including stroke \citep{li2016integrated}, cancer \citep{heidari2018prediction}, in-hospital mortality \citep{gong2017predicting}, and treatment resistance \citep{perlis2013clinical}. We consider risk stratification for patients who experience an acute coronary syndrome (ACS). An ACS entails an abrupt, reduced blood flow to the heart and refers to three types of coronary artery disease: ST-elevated myocardial infarction, non-ST-elevated myocardial infarction, and unstable angina. Treatment can range from invasive surgery to prescription drugs. Informative short-term risk metrics, can help physicians identify where patients lie on the continuum of acute coronary syndromes and make the necessary care decisions to reduce adverse outcomes. Predicting cardiovascular death (CVD) within a certain number of days of hospital admission is a common task in risk stratification literature \citep{morrow2007effects, liu2014ecg, myers2017machine}, enabling direct comparison to existing methods.
 
Risk stratification models for cardiovascular outcomes frequently operate on electrocardiogram (ECG) signals to produce risk scores \citep{acharya2003classification, rahman2015utilizing, kannathal2003classification}. Incorporating the ECG signal into a risk model is not straightforward, however, because it is a time series, often measured for days at a time. Many risk metrics compute features based on adjacent heartbeats, including the fluctuation of the ST segment \citep{klingenheben2000predictive} and the dynamic time warping distance \citep{syed2009relation}. These heuristics represent a good estimate of what a risky ECG signal may look like, but it is not well-understood what best indicates higher risk of a cardiovascular event (i.e. cardiovascular death or myocardial infarction). This presents two challenges:
\begin{enumerate}
	    \item \textit{Representation}. It is not clear what features to extract from each ECG signal. This is made especially difficult by the variable duration of heart beats.
	    \item \textit{Homogeneity}. Different levels of CVD risk at the patient level do not necessarily translate to different characteristics at the level of individual heartbeats. A high risk patient might have \textit{more} worrisome heartbeats than a low risk patient, but it is unlikely that they will have \textit{exclusively} or even a large number of worrisome heartbeats.
	\end{enumerate}
 
\paragraph{Contributions}
We  address these challenges by fusing ideas from two areas of machine learning: deep learning and Multiple Instance Learning (MIL).
	We address the representation challenge by learning predictive features directly from the data, with no task-specific engineering, using a compact neural network. This is in contrast to most existing approaches, which extract handcrafted features.
	
	To address the challenge of homogeneity, we cast the task as a MIL problem. MIL assumes that labels for collections of instances are available, but labels for individual instances are not. In our case, the instances are series of consecutive heartbeats, the collections are the instances extracted from the ECG signal for a single patient, and the label is the outcome for the associated patient. 
	
	We introduce a task-independent method for learning a signal-based risk metric. Using this method, we also present a new ECG risk score that outperforms existing ECG-based risk metrics in terms of AUC for each of 30, 60, 90, and 365 day risk of cardiovascular death.

 \paragraph{Technical Significance}
 Existing CVD risk metrics rely on hand-selected features. Outside the area of CVD, existing work in machine learning for risk stratification often relies on end-to-end learning \citep{geng2014model, farran2013predictive}. In this work, we present a risk stratification framework for ECG signals that borrows from both of these approaches, in which we learn the relationship between consecutive beats and use a simple, fixed function to aggregate this relationship. We break signal-based risk stratification into three steps: instance extraction, classification, and aggregation.  The proposed method outperforms both feature-engineered approaches and entirely learned models. 
 

\paragraph{Clinical Relevance} We aim to reduce the extent to which risk metric development depends on time-intensive feature engineering. While we validate our method only on the task of risk stratification for cardiovascular death, we make no ECG-specific choices. This suggests that our approach could also be applied to other problems.


While it is well established that the inclusion of patient features outside of the ECG increases the predictive power of a given ECG-based risk metric \citep{syed2009relation, myers2017machine}, we focus on how to incorporate the ECG signal into a risk stratification model. Using the proposed method, we predict cardiovascular death from a raw ECG signal and present a state-of-the-art risk score with respect to CVD within 30, 60, 90 and 365 days.  \\

	We include a review of related work in Section 2 and a summary of our method in Section 3. In Section 4, we outline our experimental setup and follow with results in Section 5.  We conclude with a robustness analysis in Section 6 and discussion in Section 7.

\section{Related Work}

\subsection{CVD Risk Stratification} The simplest and most commonly used means of predicting CVD is to construct a model based on easy-to-quantify patient characteristics, such as age, sex, and Left Ventricular Ejection Fraction (LVEF) \citep{cintron1993prognostic,antman2000timi, tang2007global,muntner2014validation}. There is strong evidence, however, that leveraging ECG signals can add significant predictive power \citep{liu2014ecg}. 

Because these signals take the form of long time series, it is not obvious how best to incorporate them into a predictive model. One approach is to treat the entire signal as an input to a model, either by extracting summary statistics \citep{malik1996heart} or feeding it into a recurrent neural network \citep{myers2017machine}. An alternative is to treat one ECG signal as a sequence of many examples of heartbeats. A successful means of doing so is to extract pairs of consecutive heartbeats and represent each pair using a set of informative features \citep{mccraty2015heart,syed2009relation}. These features might include polynomial fit coefficients \citep{sun2012ecg}, Legendre coefficients \citep{myers2017machine} and DTW distance \citep{liu2014ecg}. Each of these methods rely on heuristics for cardiovascular risk. In contrast, we propose a method that learns a function to estimate risk directly from the ECG signal. We build upon existing work by focusing on consecutive heartbeats and map this problem to the multiple instance learning framework.

\subsection{Multiple Instance Learning (MIL)} 
	MIL tasks involve three entities: instances, collections, and labels. In traditional supervised learning problems, each instance is associated with a label. In MIL, instances belong to collections and each collection is associated with a label \citep{amores2013multiple}.  A wealth of MIL research has occurred since the field's inception in 1998 \citep{maron1998framework} and we direct interested readers to the survey conducted by \citep{carbonneau2018multiple}.
	
	There are two basic assumptions in MIL: the standard MI assumption and the collective assumption. Our work depends upon the collective assumption, which states that positive and negative collections differ in the \textit{percentage} of instances that are positive rather than the \textit{existence} of a single positive instance. 
	
	Our approach is also an example of single instance learning \citep{foulds2010review}, in which each instance inherits the collection label. We focus on the collection label prediction task corresponding to the downstream patient outcome, as opposed to the instance label prediction task. 
	
\section{Method}
  
  We assume a collection of $M$ ECG signals $T_1,\ldots,T_M$. Each signal is associated with a unique patient $m$, and consists of $L_m$ scalar samples. Each patient is associated with a label $y_m \in \{0,1\}$, where $y_m = 1$ indicates that the patient died of cardiac death within $D$ days of hospital admission. $D$ is set to be 30, 60, 90, or 365 days after hospital adminission. Our task is to predict the label for held-out patients based on their ECG signals.  
	
	We treat this as a multiple instance learning problem. This can be formalized as the construction of three functions:
	
	\begin{enumerate}
	    \item \textbf{Instance Extractor}. A mapping $F : \mathcal{T} \rightarrow \mathcal{X}^N$, from the space of ECG signals to the space of collections of $N$ instances.
	    \item \textbf{Instance Classifier}. A mapping $G : \mathcal{X} \rightarrow [0, 1]$ from individual instances to probabilistic class predictions.
	    \item \textbf{Instance Aggregator}. A mapping $H : [0, 1]^{N} \rightarrow [0, 1]$ from predictions for each of the $N$ instances in a collection to an overall prediction for the collection.
	\end{enumerate}

	\subsection{Instance Extractor}
	
\begin{figure}
    \centering
    \includegraphics[width=\textwidth]{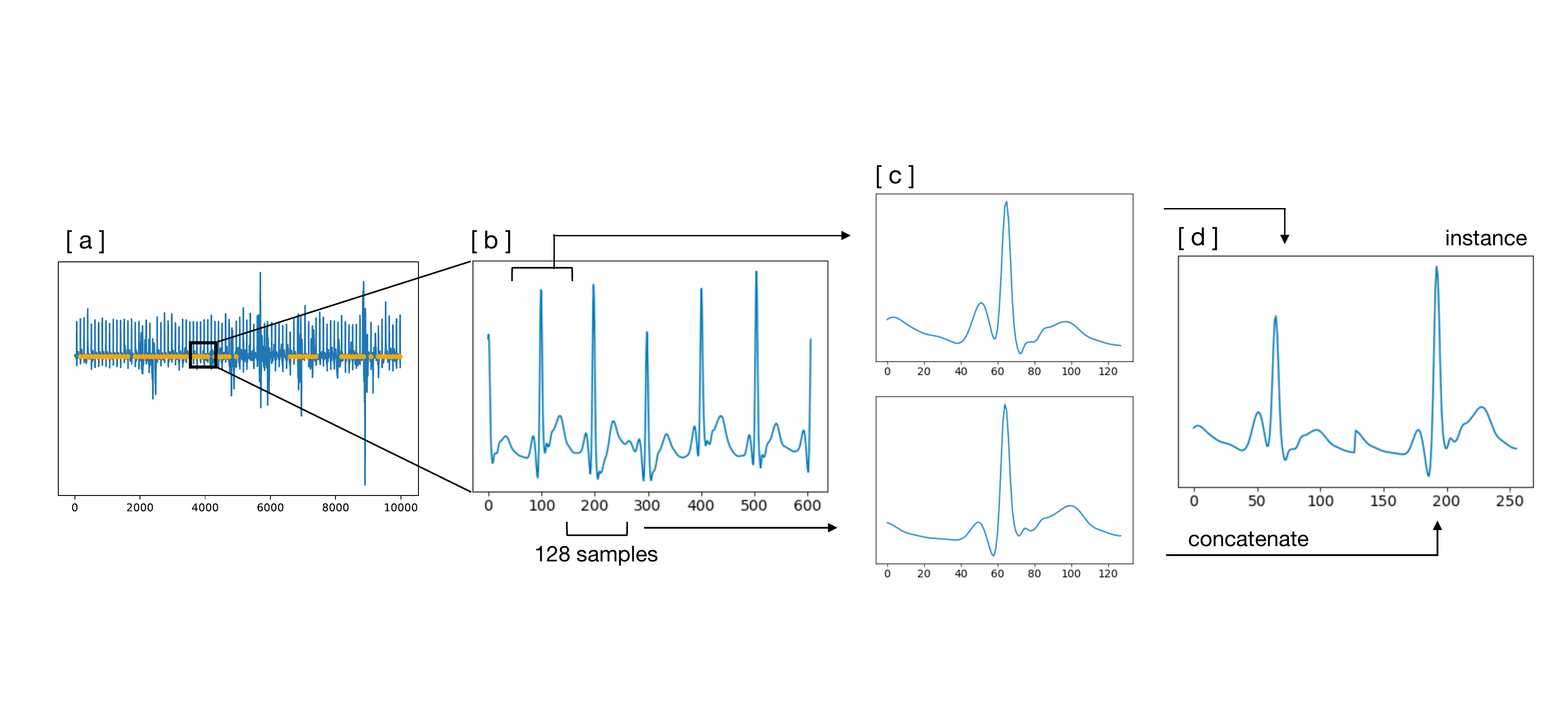}
    \caption{The four main steps of instance extraction. First, we identify non-ectopic beats(marked in orange) from the raw patient's ECG signal (a). We sample one secondc---corresponding to 128 samples---around the peak of each beat (b). This can lead to overlapping windows (c), but ensures that peaks are aligned. We then concatenate those windows (d) to create series of consecutive beats that serve as instances for the proposed method.}
    \label{construction}
\end{figure}

	 The input to the instance extractor is the patient's ECG signal, cleaned according to \ref{preprocess}. $F$ then transforms a signal $T_m$ into a set of $N_m$ instances, $X_m = \{\mathbf{x}^{i}_{m}\}_{i = 1}^{N_m}$. In the context of ECG signals, we choose instances to be groups of consecutive heartbeats. To create these groups, we identify the peak of each heartbeat using a waveform-based method \citep{martinez2004wavelet} and extract one second of data, centered around each peak.  We limit the number of instances taken from each ECG signal to 1000, corresponding to roughly 15 minutes. Our instance construction procedure is shown in Figure \ref{construction}.
	 
	 Importantly, we align the peaks of each heartbeat between instances. This is informed by previous methods that rely on transforming a patient's ECG signal into ``beat space", where adjusting for variation in heart rate between patients yields improvements in cardiovascular risk stratification \citep{liu2014ecg}. 
	 
	\subsection{Instance Classifier}

	We employ a compact convolutional neural network (CNN) to map each instance to its patient outcome. The model consists of two convolutional layers, each followed by a maxpooling layer, connected to a sigmoid output. Each convolution consists of two filters with a stride of 2 and kernel sizes of 128 and 64. The maxpooling layer has a stride of 1 and a width of 4. These sizes correspond to one second and half second windows over the ECG signal. The error in patient outcome prediction is backpropagated through the network in this step. 
    
    A CNN is well-suited to this task for several reasons. First, it extracts features directly from the raw data instead of requiring feature engineering. Second, the use of a CNN provides some translation invariance in the input space through maxpooling. This adds robustness to the behavior of the instance extraction procedure, in which overlapping windows in the instance extraction procedure produce redundant examples (see Figure \ref{construction}). Third, it can capture morphological features of signals that are indicative of downstream risk \citep{shaffer2017overview, de2004automatic}. 

	\subsection{Instance Aggregator}
	
	We aggregate instances based on the \textit{collective assumption} in MIL \citep{foulds2010review}. That is, instead of assuming that only collections with $y = 1$ have \textit{any} instances belonging to this class, we assume that collections with $y = 1$ have \textit{more} of these instances than collections with $y = 0$. We compute the probability of the collection having label $y_m = 1$ as the mean of the 20 percent of instance predictions with highest probability. I.e.,
	\begin{equation}
	    G(X_m) = \frac{1}{.2 N_m} \sum_{i = 1}^{N_m} F(\mathbf{x}^{i}_m) \cdot I[ F(\mathbf{x}^{i}_m)  \geq P_{80}( F(\mathbf{x}_m))]
	\end{equation}
	where $P_{80}(\cdot)$ is a function that computes the 80th percentile of a set of measurements. This formalizes our hypothesis that patients likely to die within $D$ days of hospital admission contain certain pathological beat sequences at a higher rate than low risk patients, but that most beats still appear normal. Following a convention common in the clinical literature, we designate patients that fall in the upper quartile of the metric as high risk and the lower three quartiles as low risk.
	
	
	\section{Experimental Set-Up}
	\subsection{Data} We use ECG recordings from the  5396 patients from the MERLIN-TIMI dataset \citep{morrow2007effects} along with downstream labels for cardiovascular death. The sampling rate is 128Hz, and each patient was recorded for approximately 48 consecutive hours upon hospital admission for ACS. We remove patients who leave the study before 90 days, leaving us with 5245 patients. We also exclude 270 patients with missing patient attributes, to enable  fair comparison to existing methods. Of the remaining 4975 patients, 107 patients died of CVD within 90 days and 217 died of CVD within 365 days. While there is an extreme class imbalance for all time horizons, this imbalance is particularly severe at short time scales.

	Demographic factors for CVD and non-CVD patients are similar, as are their distributions of known correlates (including age and smoking status). The dataset is further divided into 5 randomized train/test splits. We select our method's hyperparameters (such as instance length and prediction function percentile) using one split, and test on the remaining four splits.
	
	\subsection{Implementation}
	
	We implement the network in Keras \citep{chollet2015keras} and TensorFlow \citep{abadi2016tensorflow}, and optimize it using Adam \citep{kingma2014adam} with default parameter settings for learning rate and decay. To address class imbalance, we balance each batch for positive and negative examples. This leads to over sampling the positive class.

	\subsection{Preprocessing}
	\label{preprocess}
	 Three pre-processing steps are typical when working with ECG data. The first is to remove ectopic beats from the ECG signal. Next is to remove baseline wander, or noise in the signal caused by patient motion or respiration. Finally, we normalize the entire signal based on each patient's R-wave amplitude in order to correct for inter-patient differences in measurement. This protocol is described by \citep{liu2014ecg}, and we perform the relevant steps using the same Physionet Signal Quality Index package \citep{li2007robust}.

	\subsection{Baselines} We evaluate the proposed approach against two sets of benchmarks: existing CVD risk metrics and variations of the proposed method. Compared to existing work in multiple instance learning, the instance number and instance dimension of our dataset far exceeds the typical scale of MIL datasets \citep{cheplygina2015characterizing}, disqualifying several kernel-based methods because of storage constraints. The only general-purpose MIL method able to run on our dataset was STK \citep{gartner2002multi}. STK applies a specialized SVM, which defines collection similarity as the dot product of collection-level statistics including mean, variance, and standard deviation \citep{gartner2002multi}. This method consistently yielded AUCs near 0.5 on our data so we omit comparison to it.
    
	We evaluate against three existing CVD risk metrics: TRS \citep{antman2000timi}, MVB \citep{liu2014ecg}, and LR-RNN \citep{myers2017machine}. Morphological variability in beat-space (MVB) measures risk for CVD by averaging the dynamic time warping distance between adjacent beats. The method then learns the best frequency at which to compute variability between adjacent beats. LR-RNN supplements the first minute of a patient's ECG signal with commonly used patient features. 
	
	We also test against three variations of our proposed method: MIL-LR, MIL-FCX, and MIL-Set. MIL-LR uses a logistic regression in place of the the CNN, while MIL-FCX employs a one-layer fully connected network consisting of X hidden units where $X \in {2, 3}$. We include these as baselines to evaluate the use of a neural network and, in particular, the use of a CNN. MIL-Set uses a CNN for the instance classification step but learns the aggregation function, using an approach similar to \citep{zaheer2017deep}. 

\section{Results} 

In this section, we evaluate our method's ability to identify patients at risk for CVD within 30, 60, 90 and 365 days of hospital admission for an acute coronary syndrome. We measure performance using the area under the receiver operating curve (AUC) and the odds ratio.

\subsection{Prediction Horizon}

We present the performance of the proposed model, MIL-CNN, in terms of AUC in Table \ref{aucs}. The method achieves state-of-the-art AUCs across all time horizons examined. MIL-CNN achieves at least a 10 point increase in AUC over each horizon compared to MVB. MIL-LR performs comparably to MVB, while MIL-FC2 performs better. This suggests the value of non-linearities in learning a representation over instances. Using a CNN as the instance classifier offers further improvement, surpassing LR-RNN in terms of mean AUC. MIL-Set performs poorly, implying that the additional task of learning the aggregation function hinders the model. LR-RNN is within a standard deviation of the proposed model's performance, but also makes use of patient data external to the ECG signal. We also compare MIL-CNN to the same CNN architecture trained across the entire signal (eliminating the instance segmentation step) and show that MIL-CNN achieves higher AUCs. The table with these results can be found in Appendix A.
     For the existing CVD risk metrics, we see a steady increase in AUC as we go from the 30 day horizon to the 90 day horizon, followed by a small dip between the 90 day and 365 day prediction performance. This dip could be because symptoms of cardiovascular death at longer time scales---between 90 and 365 days---pose a more challenging identification task than symptoms of cardiovascular death within 90 days. 

\begin{table}
\centering
\resizebox{\linewidth}{!}{%
    \begin{tabular}{c|ccc|ccc|c}
    \hline
          Days & TRS & MVB & LR-RNN & MIL-LR & MIL-FC2 & MIL-CNN & MIL-Set   \\
    \hline
         30  & $0.68 \pm 0.067$ & $0.66 \pm .055$ & $0.78 \pm 0.043$ & $0.67 \pm .022$ & $0.71 \pm .020$ & $\mathbf{.83 \pm .007}$ &  $0.54 \pm .076$ \\
         60  & $0.69 \pm 0.071$ & $0.68 \pm .046$ & $0.78 \pm 0.050$ & $0.68 \pm .017$ & $0.75 \pm .007$ & $\mathbf{.79 \pm .018}$ &  $0.54 \pm .075$\\
         90  & $0.70 \pm 0.073$ & $0.68 \pm .047$ & $0.79 \pm 0.052$ & $0.71 \pm .031$ & $0.78 \pm .009$ & $\mathbf{.81 \pm .003}$ &  $0.64 \pm .080$\\
         365 & $0.70 \pm 0.055$ & $0.66 \pm .060$ &   $0.74 \pm 0.040$ & $0.70 \pm .025$ & $0.75 \pm .014$ & $\mathbf{.78 \pm .005}$ & $0.68 \pm .032$
    \end{tabular}}
    \caption{We report the AUC and standard deviation over four splits of the MIL framework compared to existing CVD risk metrics (TRS, MVB, and LR-RNN). Within the MIL framework, we test across three instance classifiers: LR, FC2, and a CNN. We also test against an end-to-end learning approach, MIL-Set. We see that across all time horizons, MIL-CNN outperforms existing methods. }
    \label{aucs}
\end{table}

\subsection{Odds Ratios}

A quarter of the test patients are designated as high risk. This amounts to 312 high risk patients and 935 low risk patients. We compare the ORs obtained by our method to MVB, the only one of our baselines that operates exclusively on the ECG signal. Across all time horizons, our method achieves a higher average OR across splits. The greatest improvement is seen at the 90 day time horizon, in which MIL-CNN obtains a median OR of 6.33 and identifies 19 of 28 patients who suffer from cardiovascular death as high risk. 
\\

\begin{figure}[!t]
    \centering
    \includegraphics[width=.7\textwidth]{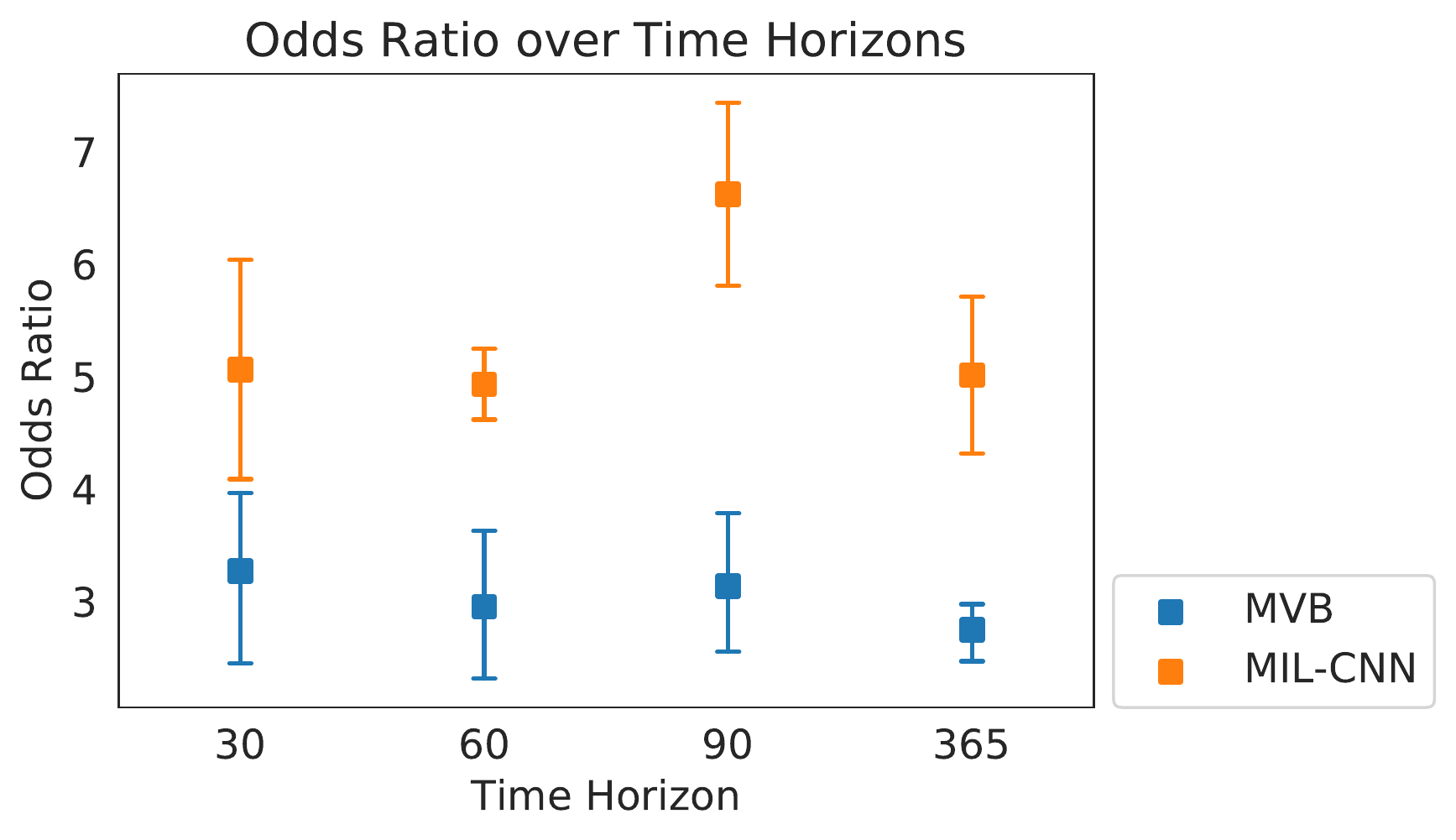}
    \caption{Distribution of odds ratios over time horizon tasks. We consider only MVB and MIL-CNN, the two risk metrics that operate on the patient ECG signal. We report the average and standard deviation of the odds ratios over 5 splits. MVB maintains an odds ratio of about 3 at each time horizon. MIL-CNN demonstrates a higher odds ratio when forecasting for 90 days, but otherwise obtains an odds ratio of around 5.}
    \label{or_scores}
\end{figure}

\begin{figure}[t]
\centering
Instance Length vs. AUC \quad \quad\quad\quad\quad\quad\quad\quad\\
\resizebox{\linewidth}{!}{%
\begin{tabular}{ccc}
\includegraphics[width=65mm]{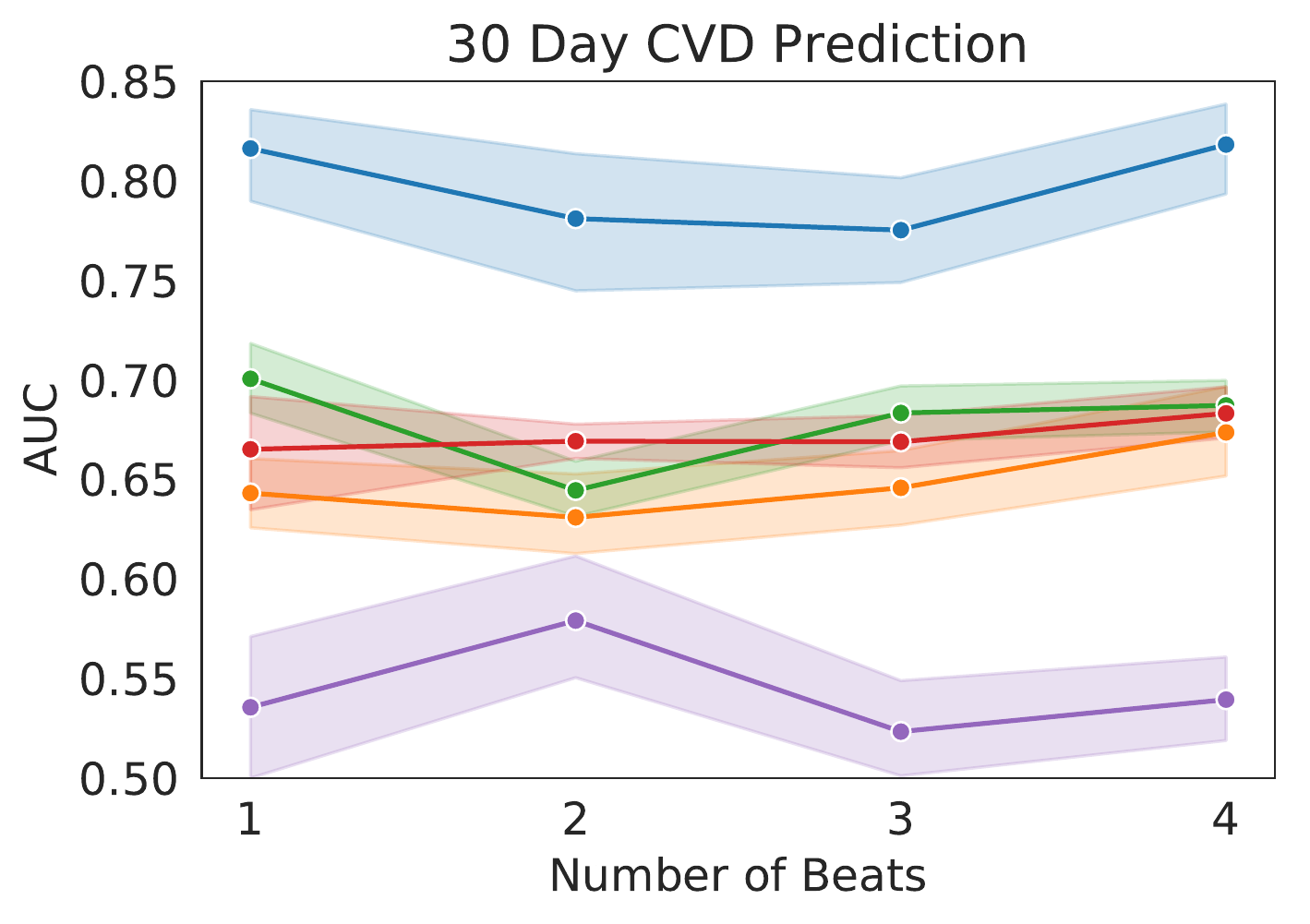} &   \includegraphics[width=65mm]{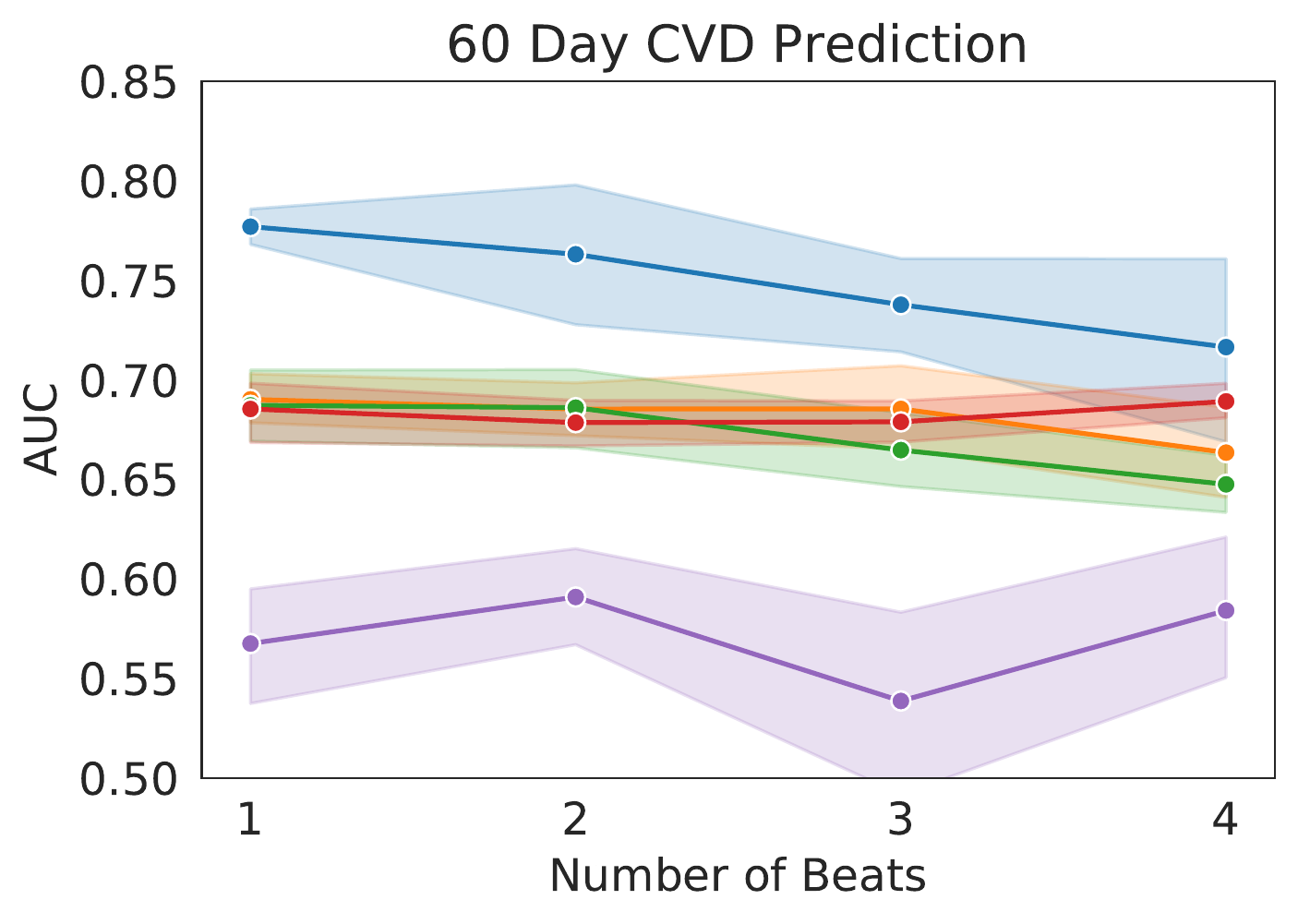} & \\
 \includegraphics[width=65mm]{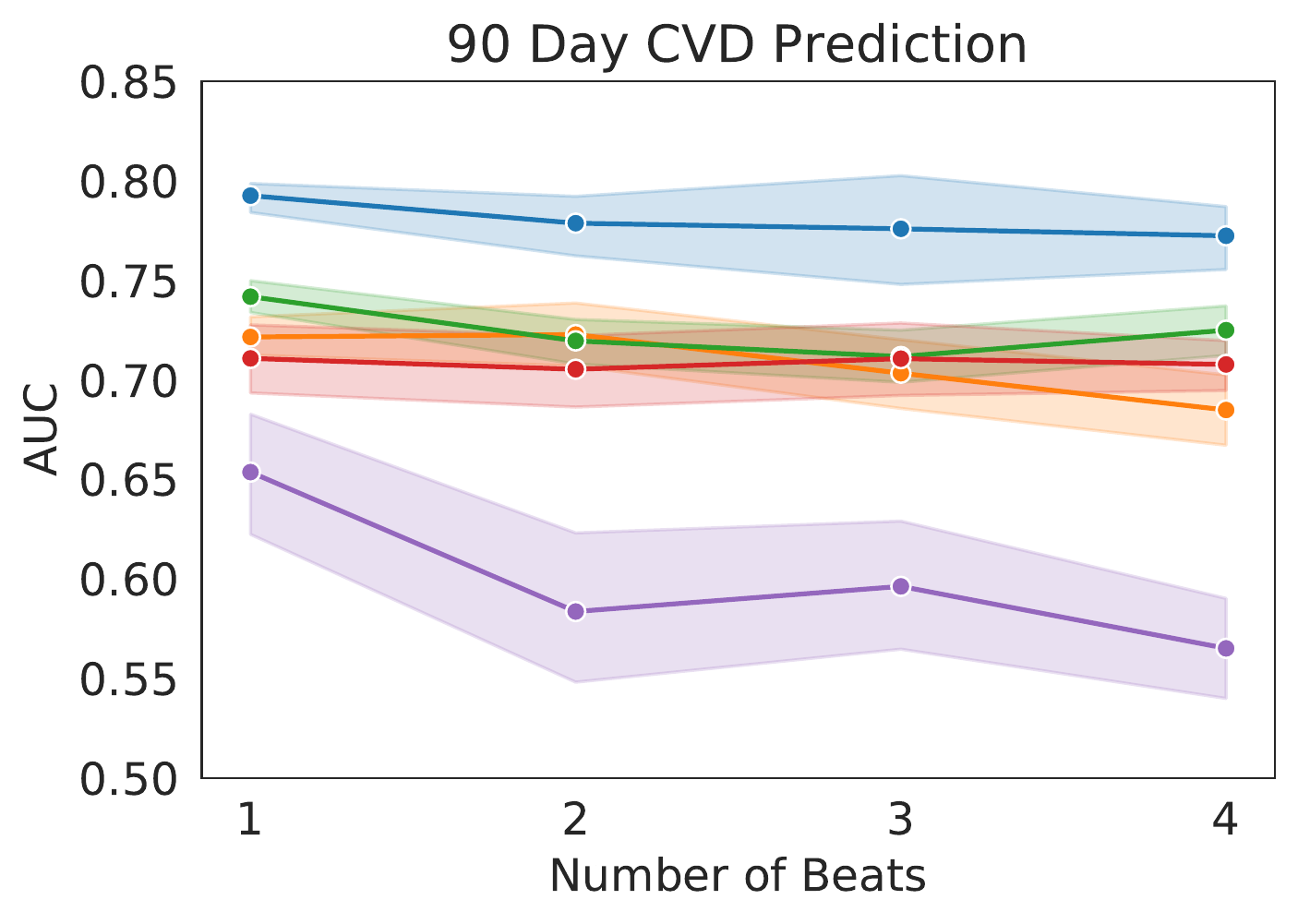} &   \includegraphics[width=65mm]{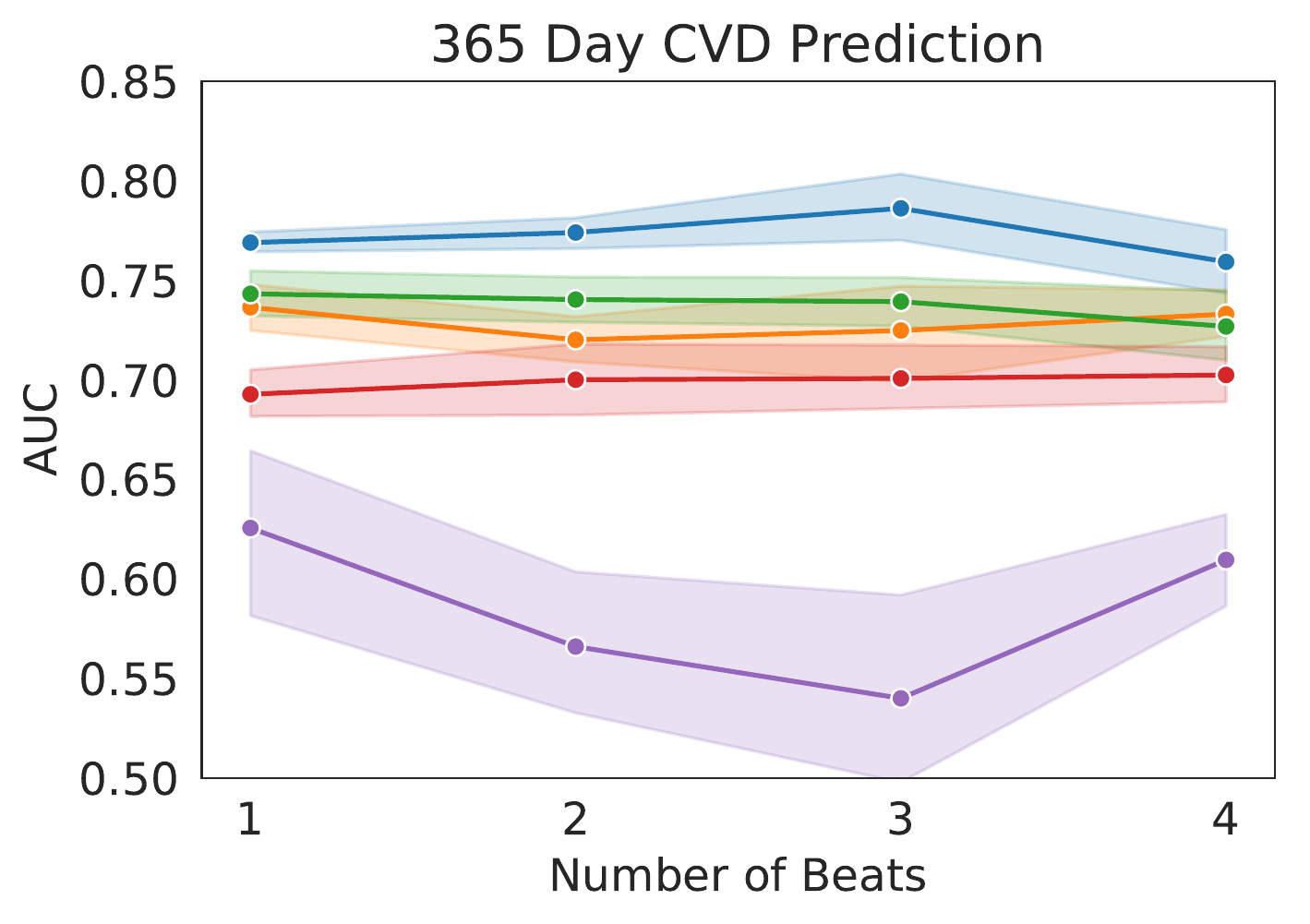} & 
 \includegraphics[width=20mm]{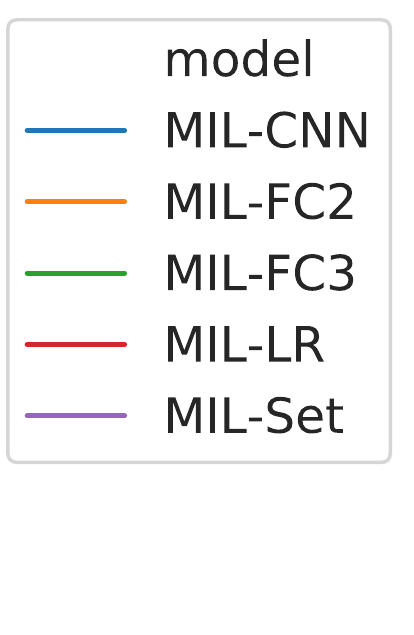}
\end{tabular}}
\caption{Effect of number of beats per instance on AUC at different time horizons. At the top, we plot the MIL method's performance with a CNN as the instance classifier. Lines clustered in the middle represent MIL performance with variations on the instance classifier. The purple line shows MIL-Set's performance on the task. The plots demonstrate a resilience to instance choice superior to that of MIL-Set, across all horizons.}
\end{figure}

\section{Robustness}

We include tests to demonstrate our method's robustness to both the choice of instance extraction function and the choice of instance aggregation function. We also show that our method is more resilient to class imbalance than MIL-Set. We include MIL-Set to show the benefit a simple, fixed aggregation function offers. 

\subsection{Robustness to Instance Extraction Function}

In this section, we evaluate the sensitivity of MIL-CNN to the instance extraction function. Segmentation into consecutive heartbeats is well-justified based on existing literature and the intrinsic periodicity of the signal, but the optimal number of consecutive heartbeats is not obvious. We test each MIL method across four instance types: one, two, three, and four consecutive heartbeats. Our method is robust to this choice.

\subsection{Robustness to Aggregation Function}

\begin{figure}
    \centering
    \includegraphics[width=.75\textwidth]{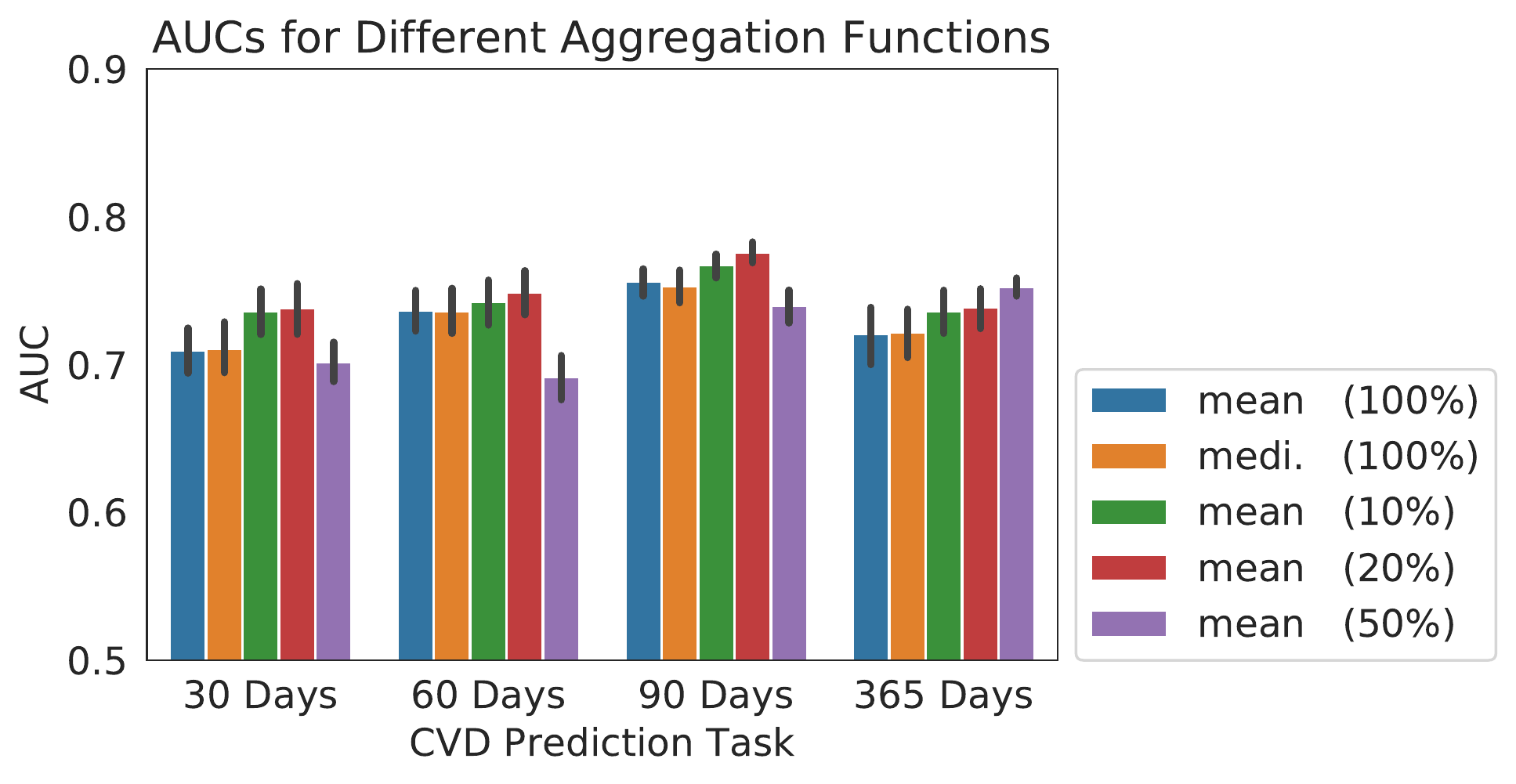}
    \caption{Performance of aggregation functions across a variety of horizons. We see a minor increase in AUC when using the mean of the top 20 percentile of instance predictions on all horizons except 365 days. Each AUC is averaged across 5 splits and instance classifiers. }
    \label{agg_fs}
\end{figure}

We also test alternatives to averaging the top 20\% of the instance predictions. The motivation for this initial choice lies in how every ECG signal segment need not---and often does not---present information relevant to patient risk for a particular outcome. First, we replace the mean with the median. We assess the mean of the top 10\%, 20\% and 50\% of the instance predictions. We show our results in Figure \ref{agg_fs}, demonstrating that the choice of aggregation function--among those tested--has no significant effect on the model's AUC. 


\subsection{Robustness to Class Imbalance}

\begin{figure}[t]
    \centering
    \includegraphics[width=.65\textwidth]{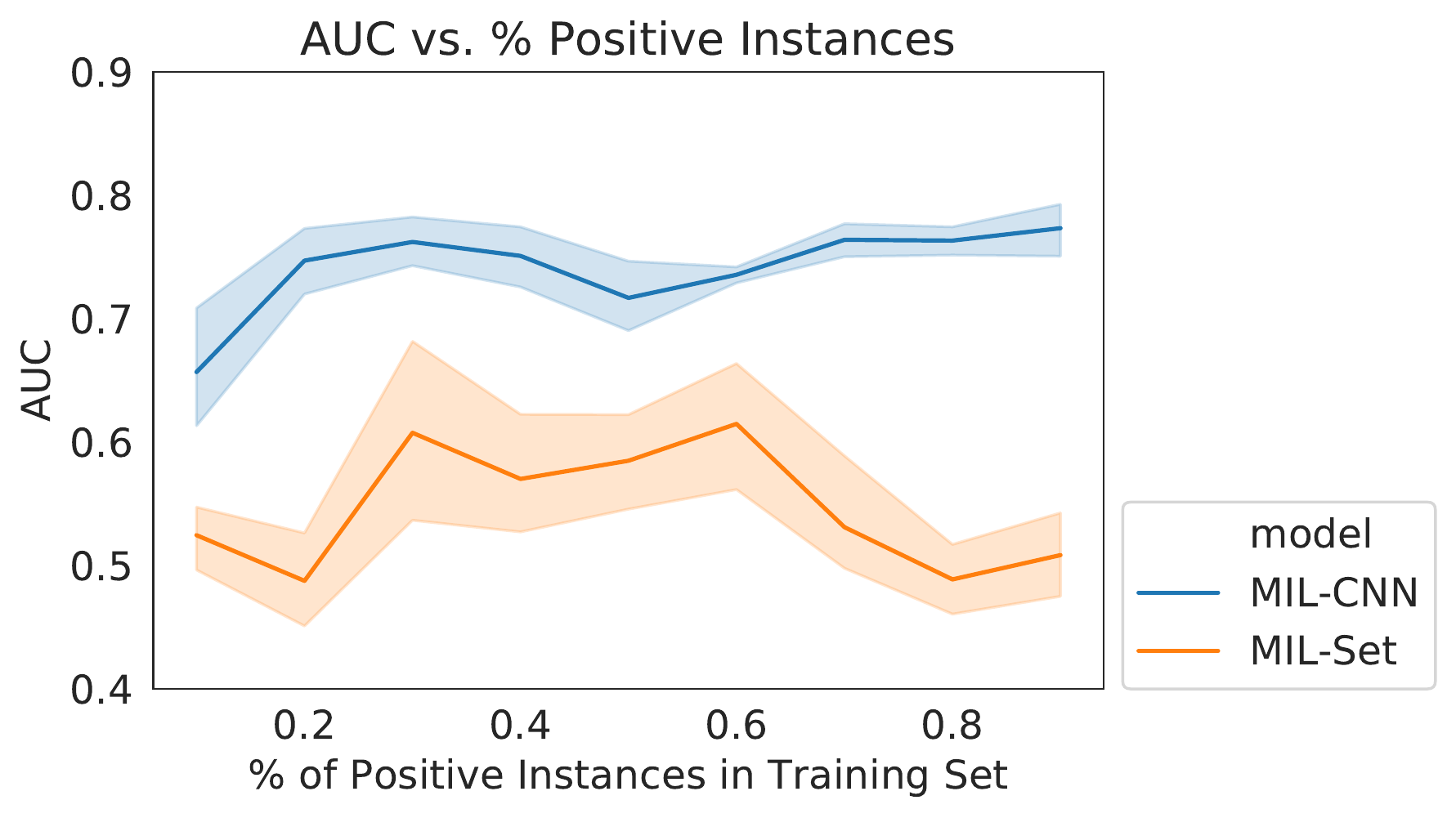}
    \caption{Testing for model robustness to the number of positive training examples on the 90-day risk stratification task. The x-axis is the percentage of existing positive instances provided to the model during training. The model normally trains on 77 positive examples and 3651 negative examples. This plot demonstrates two trends: 1) MIL-CNN is not sensitive to extreme class imbalance, learning a reasonable risk metric with only 15 positive examples and 2) MIL-CNN demonstrates less variance in its performance over multiple runs on the same split.}
    \label{npos}
\end{figure}

 We vary the number of positive instances in each training set---providing between 10-90\%  of the available positively labeled training data and measure the resulting AUCs. As shown in Figure \ref{npos}, the proposed model remains resilient to extreme class imbalance while MIL-Set does not. As the number of positive instances increases, there is no clear trend with MIL-Set performance.

\section{Discussion} 

In this work, we propose a successful way to incorporate a patient's electrocardiogram signal into a risk stratification model. We validate our approach by building a risk metric to identify patients at high risk for cardiovascular death after an acute coronary syndrome and report state-of-the-art AUCs. Moreover, we present these results for four time horizons: 30, 60, 90, and 365 days. 
 
We also show that our method is robust to hyperparameter choice and continues to excel in settings that demonstrate severe class imbalance. Two aspects of the method are  worth highlighting. First, we perform no feature engineering, which suggests the method's promise in other signal-based risk stratification scenarios. Second, we rely exclusively on a patient's ECG signal. This suggests that we could obtain an even better risk metric with the addition of patient-specific features.

\section{Future Work} 

The success of our approach suggests at least three directions for future work. First, our ability to forecast long-term outcomes from ECG signals suggests that models for risk stratification at longer time scales---on the order of years instead of months---may be worth exploring.  This has remained a relatively understudied problem \citep{bueno2016long}, but may be tractable using extensions of the ideas presented here. Second, because there is little in our approach that is specific to ECG signals, it could be applied to risk stratification using other biometric signals as well. Third, while the objective of this task is binary classification, multiple instance regression could allow us to forecast patient outcomes, such as day of cardiovascular death, at a more granular level.

\acks{We would like to thank Paul Myers, Jen Gong, and Yun Liu for their helpful feedback. This work was funded by the National Science Foundation Grant 1122374.}
\bibliography{sample}
\pagebreak

\appendix 

\section{CNN Comparison}
\begin{table}[H]
\centering
\resizebox{.4\linewidth}{!}{%
    \begin{tabular}{c|ccc}
    \hline
          Days & TRS & MIL-CNN   \\
    \hline
         30  & $0.68 \pm 0.067$ & $0.66 \pm .055$ \\
         60  & $0.69 \pm 0.071$ & $0.68 \pm .046$ \\
         90  & $0.70 \pm 0.073$ & $0.68 \pm .047$ \\
         365 & $0.70 \pm 0.055$ & $0.66 \pm .060$  
    \end{tabular}}
    \caption{We compare MIL-CNN to a CNN trained on the contiguous, unsegmented signal. We see that MIL-CNN outperforms the plain CNN, indicating the importance of the multiple instance learning framework.}
    \label{cnn_aucs}
\end{table}

\end{document}